\documentclass[letterpaper, 10 pt, conference]{ieeeconf}

\IEEEoverridecommandlockouts

\usepackage{graphics}
\usepackage{amsmath}
\usepackage{amssymb}
\usepackage{todonotes}

\title{\LARGE \bf
Fusing multimodal neuroimaging data with a variational autoencoder
}

\author{Eloy Geenjaar$^{1,2}$, Noah Lewis$^{1}$, Zening Fu$^{1}$,  Rohan Venkatdas$^{1,3}$, Sergey Plis$^{1}$, Vince Calhoun$^{1}$
\thanks{*This work was not supported by any organization}
\thanks{$^{1}$ Tri-institutional Center for Translational Research in Neuroimaging and Data Science (TReNDS), Georgia State, Georgia Tech, Emory,
        Atlanta, GA 30303, USA
        {\tt\small e.geenjaar@gsu.edu}}%
\thanks{$^{2}$ Faculty of Electrical Engineering, Mathematics \& Computer Science, TU Delft, Delft, the Netherlands}%
\thanks{$^{3}$ Lambert High School, Suwanee, GA, USA}%
}

\begin{document}

\maketitle
\thispagestyle{empty}
\pagestyle{empty}

\begin{abstract}
Neuroimaging studies often involve the collection of multiple data modalities. These modalities contain both shared and mutually exclusive information about the brain. This work aims at finding a scalable and interpretable method to fuse the information of multiple neuroimaging modalities using a variational autoencoder (VAE). To provide an initial assessment, this work evaluates the representations that are learned using a  schizophrenia classification task. A support vector machine trained on the representations achieves an area under the curve for the classifier's receiver operating characteristic (ROC-AUC) of $0.8610$.
--
\newline

\indent \textit{Clinical relevance} -- This work helps examine the complex interplay between multiple neuroimaging modalities and how that interplay affects mental disorders.
\end{abstract}

\section{INTRODUCTION}

Multimodal neuroimaging data is abundantly available and although approaches that seek to combine these data, e.g., JointICA \cite{jointica}, and more recently multimodal subspace analysis \cite{silvamultimodalsubspace} focus on linear decompositions, recent work on multimodal deep learning offers the benefits of additional flexibility which can also capture nonlinear relationships. Multimodal deep learning research mostly focuses on the relationship between audio, images, and/or text \cite{multimodaldl}. The exciting new direction of multimodal representation learning, together with growing evidence that deep learning representations can provide robust biomarkers \cite{aneesnature}, paves the way for multimodal representation learning in neuroimaging. 

Just as with a puzzle, pieces that fit together will contain both shared information and mutually exclusive information. Fusing these pieces in terms of neuroimaging can lead to biomarkers that more robustly predict changes associated with mental illnesses \cite{multimodalcalhoun}. An important downside to deep learning techniques is that their non-linear nature can present challenges to interpretation, which undermines their applicability to medical problems. Interpretability is, therefore, an important consideration in this work.

Recent work in multimodal deep learning applied to neuroimaging has focused on information maximization between representations extracted from two modalities \cite{afedorovgeneralization, afedorovtaxonomy} or by translating between modalities \cite{multimodalplis}.  This work aims to learn a continuous manifold of multiple modalities so that they are represented in a locally Euclidean space. The model architecture that is used is a variational autoencoder (VAE) \cite{vae}, which maximizes a lower bound on the log-likelihood of the data's marginal distribution. Other work on multimodal VAEs focuses on a factorization of shared and private subspaces \cite{mmvae} and uses a separate encoder for each modality. In this work, we intentionally force all of the modalities to inhabit the same shared subspace by using a single encoder-decoder pair for all modalities. This allows us to interpolate between modalities, similar to how VAEs have previously been used to interpolate between different digits in the MNIST dataset \cite{vae}.

To provide an initial assessment of the potential for the unsupervised training of the VAE to produce robust biomarkers for complex mental illnesses, we evaluate our model and its representations on a schizophrenia classification task. Schizophrenia is a mental illness that is characterized by complex interconnected changes in dynamics and functional connectivity. To understand how the brains of patients with schizophrenia differ from controls it is imperative to piece together information from multiple modalities \cite{multimodalcalhoun}, such as structural MRI and functional MRI. In this work, we treat a structural MRI (sMRI) volume and each of the intrinsic functional brain networks that are extracted from resting-state functional MRI (rs-fMRI) data using NeuroMark \cite{neuromark} as separate modalities. An important consideration when choosing our method was that a VAE can decode locations in its latent space to brain space and provide insight as to what regions in the brain may differ between two groups. The regions that have in previous literature been linked to schizophrenia include the thalamus, cerebellum, caudate, superior temporal gyrus, most of the visual system (e.g.,  lingual gyrus, occipital gyrus \cite{icasz}), and the supplementary motor area \cite{smasz}.

\section{Contributions}
This work introduces an interpretable approach for fusing multiple neuroimaging modalities with the following properties:
\begin{itemize}
    \item It focuses on fusing multiple modalities and it scales in its number of parameters as $\mathcal{O}(1)$ with the number of modalities.
    \item The model optimizes both an encoder and a decoder, the decoder makes it easier to interpret group differences in the latent space, because locations can be decoded back into brain space.
    \item The framework forces the modalities to be represented in a shared locally Euclidean space, instead of learning two spaces and maximizing their similarity. This allows us to interpolate directly between multiple modalities.
\end{itemize}

\section{Method}

\subsection{Problem setting}

Let $\{M_i = \{x_{i,j},...,x_{i, N}\}\}_{i=1,...,n}$ be a set of modalities with $N$ samples per modality and $n$ modalities. Sample index $j$ corresponds to a subject and each subject will have a sample for each modality. Instead of learning each modality with a separate decoder, we enforce a shared subspace. Further, to make this approach scalable to a large number of modalities and because we already use a shared decoder, we also only use one encoder for all modalities. This forces the features that are learned for each modality to be similar and makes sure the model scales well in terms of memory usage. Further, given that neuroimaging datasets are considered small compared to more commonly used deep learning datasets, using multiple encoders may lead to overfitting. The encoder decoder couple is optimized with respect to the log-likelihood of the marginal distribution of $M$ and each volume is treated as an independent sample. The integral over the marginal distribution of $M$, $p_{\theta}(x) = \int p_{\theta}(z) p_{\theta}(x|z) \mathit{d}z$ is intractable. We therefore optimize the following evidence lower bound (ELBO) using a variational autoencoder (VAE) \cite{vae}:

\begin{equation}
    \label{eq:elbo}
    \begin{aligned}
    \log p_{\theta}(x_{i, j}) &\geq \mathcal{L}(\theta, \phi; x_{i, j}) \\
    &= \mathbb{E}_{q_{\phi}(z|x)} \left [- \log q_{\phi} (z|x) + \log p_{\theta}(x, z) \right] \\
    &= -D_{\text{KL}}(q_{\phi}(z|x_{i,j}) \ || \ p_{\theta}(z)) \\ & \quad +    \mathbb{E}_{q_{\phi}(z|x_{i,j)}} \left[\log p_{\theta}(x_{i,j}|z) \right]
    \end{aligned}
\end{equation}

The objective function is calculated as a sum over each data point and can be split up into three parts, the first part is an encoder $q_{\phi}(z|x_{i,j})$ parameterized as a convolutional neural network (CNN) with parameters $\phi$ that estimates the latent variable $z$. The second part is a decoder $p_{\theta}(x_{i,j}|z)$, also parameterized as a CNN with parameters $\theta$, that reconstructs the original sample $x_{i,j}$ from the estimated latent variable $z$. The last part of the loss is the KL-divergence between a prior of our choosing $p_{\theta}(z)$, which we choose to be a diagonal multivariate Gaussian centered at $0$ with a standard deviation of $1$, and the estimated latent variable $z$. The latent variable $z$ is sampled from a multivariate Gaussian as well, which in turn is parameterized by a mean $\mu$ and variance $\sigma$ that is estimated by the encoder.

\subsection{Classification}
To evaluate whether the fusion of modalities in the VAE's latent space leads to robust biomarkers, we set up a classification task. The model is first trained using 10-fold cross-validation, where each fold of subjects is used as a test set once and the other 9 folds are used to train on. The validation set is randomly selected as a stratified 10\% of the subjects in the training set. After training the VAE in an unsupervised manner, the weights in the VAE are frozen. The complete dataset is then embedded using the encoder $q_{\phi}(z|x_{i,j})$, where instead of sampling $z$ from its estimated multivariate Gaussian, we use the estimated mean $\mu$ as our latent variable $z$. This is to make sure there is no stochasticity in the inference process. The representations of the training and validation sets are stacked and used as input for a machine learning model, this model is then evaluated using the test set representations, as follows.

The estimated latent variable $z$ can be interpreted as a low-dimensional representation of a volume, with a dimensionality $l$. Given that each subject has $n$ different modalities, each subject will also have $n$ representations $z_1,...,n$. These representations can be concatenated for a subject to create a feature vector with a size of $n \times l$. The subject-by-feature matrix can be used as input for a classifier. In this case, we train a support vector machine (SVM) 
 
to predict whether subjects in a held-out test set are patients with schizophrenia. Given that each modality is represented using $l$ features, we can extract the feature importance for all $nl$ features and then sum the features for each modality, to get feature importance for each of the $n$ modalities. The feature importance information helps us understand how brain changes related to schizophrenia are jointly represented in multiple modalities. 

The SVM classifier is evaluated by calculating the area under the curve (AUC) of their receiver operator characteristic (ROC). Given that the VAE is trained in an unsupervised manner and that parts of the training process are stochastic, we evaluate the model for $5$ different seeds to make sure the method is robust to its seed, these experiments are performed with a latent dimensionality of $128$. To evaluate the effect of the number of latent dimensions on the classification performance, we set the seed to be $42$ and trained the model with four different latent dimensionalities $128$, $256$, $512$, $1024$. We select the weights obtained during the first training fold for the model that performs the best. The best performance is determined by averaging the ROC-AUC over the 10 folds for each model. These encoder and decoder weights are used to create the figures and to determine the importance of each modality for the classification of schizophrenia.

\subsection{Data}
The datasets used in this study are FBIRN, B-SNIP, and COBRE, each dataset was processed using NeuroMark \cite{neuromark} to obtain 53 independent component networks (ICNs). These 53 ICNs, together with a structural MRI scan for each subject are considered to be separate modalities, so $n = 54$. The sMRI data is preprocessed using SPM 12 in a Matlab 2016 environment. The data is then segmented into gray matter volume (GMV) with the help of a modulated algorithm, the GMV is then smoothed with a $6$mm FWHM Gaussian kernel.
\newline
Each ICN is a volume with $53\text{-by-}63\text{-by-}52$ voxels, the sMRI volumes are resized to that same size using Scipy \cite{scipy}. The values in each volume are then rescaled to [-1, 1] by dividing the values in a volume by their maximum, which is also sometimes referred to as maximum absolute scaling. The dataloader and transformations were implemented with the help of TorchIO \cite{torchio}. 

\subsection{Implementation}
The batches are constructed by loading the $53$ ICN volumes and an sMRI volume for a subject and concatenating them into a batch. Each volume is considered to be sampled independently during training, the reason the volumes are loaded per subject however is to minimize disk accesses. The ICNs for one subject are all saved in one file, so loading all of them into a batch leads to a smaller number of disk accesses and reduces training time. Further, because each modality is equally present in a mini-batch for which an optimizer step is performed, the loss calculated over each batch is balanced for the modalities.

The code for the model, inference, and training was implemented using PyTorch \cite{pytorch}, Catalyst \cite{catalyst}, and NumPy \cite{numpy}. The VAE uses a convolutional encoder and decoder pair, each of the layers uses a $3$-voxel kernel, a stride of $2$ and $1$-voxel padding. The channel sizes in the encoder are $1\to64, 64\to128, 128\to256, 256\to512$ and $512\to256, 256\to128, 128\to64, 64\to32, 32\to16, 16\to1$ in the decoder, the last layer in the decoder uses stride $1$ and no bias parameters. Each convolutional layer uses a ReLU \cite{relu} as its activation function, except for the last layer in the decoder, which uses a hyperbolic tangent function to map the output between [-1, 1] to match the input range. The last convolutional layer in the encoder produces an output with shape: $4\text{-by-}4\text{-by-}4$ and $256$ channels, this output is flattened and mapped to the mean $\mu$ and variance $\sigma$, which are used to construct a diagonal multivariate Gaussian from which $z$ is sampled. To make sure the VAE is fully differentiable, we use the reparameterization trick to train it \cite{vae}. The classification evaluations in the latent space are implemented using RAPIDS AI \cite{rapidsai} to make sure highly parallelizable computations are performed on the GPU and to minimize costly CPU$\to$GPU and GPU$\to$CPU data transfers. The experiments were performed on an NVIDIA DGX-1 V100.

\subsection{Latent structure}
\begin{figure}[h]
    \centering
    \includegraphics[width=0.62\linewidth]{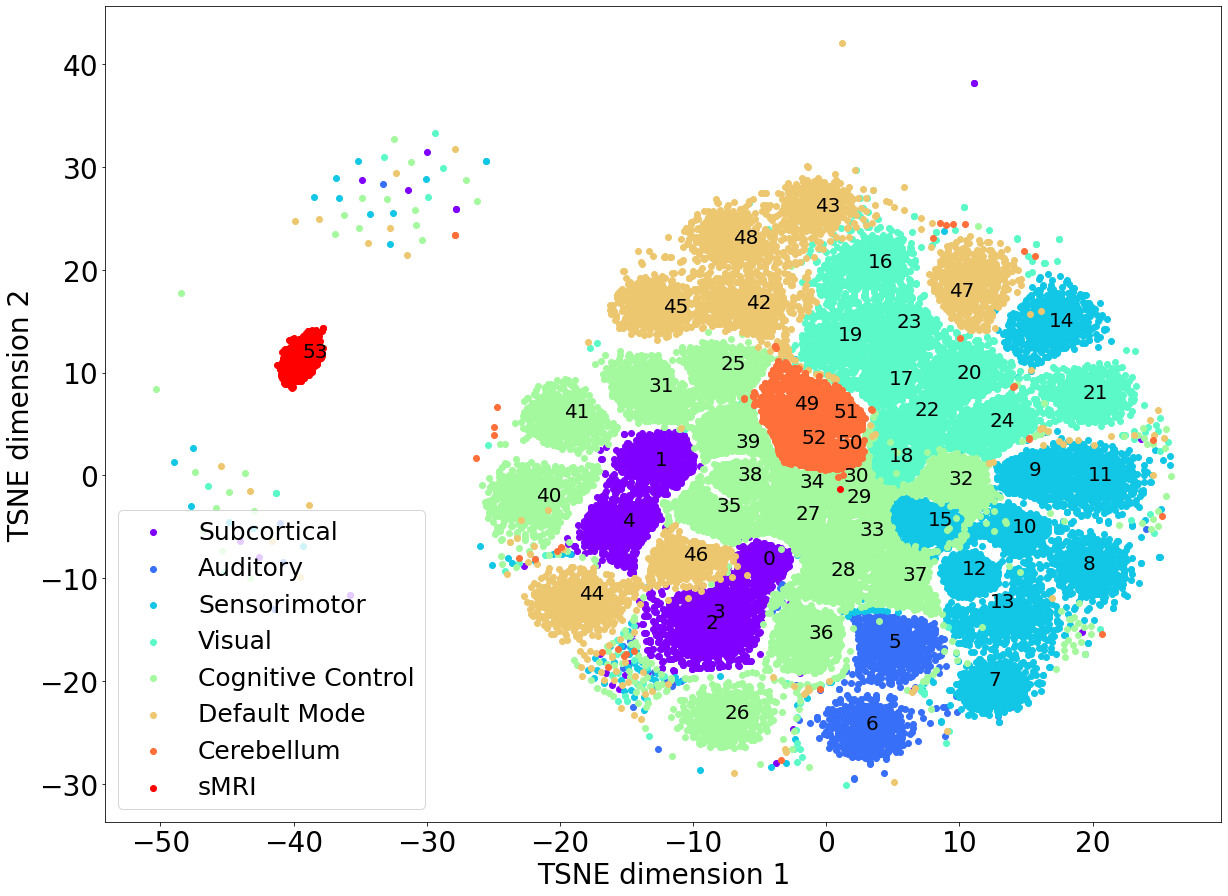}
    \caption{A T-SNE projection of the modality clusters in the latent space, each number indicates a different modality starting at 0. Each color indicates the domain that the intrinsic networks belong to. Each subject is represented by $54$ points in this plot, one for each modality.}
    \label{fig:clusterlatentspace}
\end{figure}

Most of the modalities that are used in this paper are intrinsic networks, which are obtained through independent component analysis (ICA). The independence in the spatial volumes for those components leads to a latent space that clusters modalities, which is shown in Figure \ref{fig:clusterlatentspace}. The plot depicts a T-SNE projection of a $512$-dimensional space that preserves the latent structure in a 2D image. Interestingly, the ICNs that belong to the same domain are generally clustered together, such as ICNs in the cerebellum. It is also clear from Figure \ref{fig:clusterlatentspace} that the sMRI cluster is located relatively farther away from the other modalities in the latent space. The ICNs are all localized spatial ICA maps, whereas the sMRI volume represents all of the structures in the brain. There is more inter-subject variance to be modeled for the sMRI volumes than for the spatially localized ICNs which likely contributes to the sMRI cluster being further away from the latent space ICN clusters.

\section{Results}

\subsection{Classification}

The average ROC-AUC for the five models trained with a latent dimensionality of $128$ and multiple different seeds is $0.8374$, with a standard deviation of $0.0026$. This shows that the model robustly learns a latent space, where patients with schizophrenia and controls are linearly separable.

The experiment involving an increasing number of latent dimensions shows that the ROC-AUC increases with the number of latent dimensions up to a certain point. The ROC-AUC is $0.8354$ for $128$ dimensions, $0.8569$ for $256$ dimensions, $0.8610$ for $512$ dimensions and $0.8540$ for $1024$ dimensions. These results show that the model is not learning to represent each modality using a single latent dimension, but rather that the variance that is modeled across a latent dimension is shared among multiple modalities or that multiple latent dimensions are used to model a single modality.  

\begin{figure}[h]
    \centering
    \includegraphics[width=0.8\linewidth]{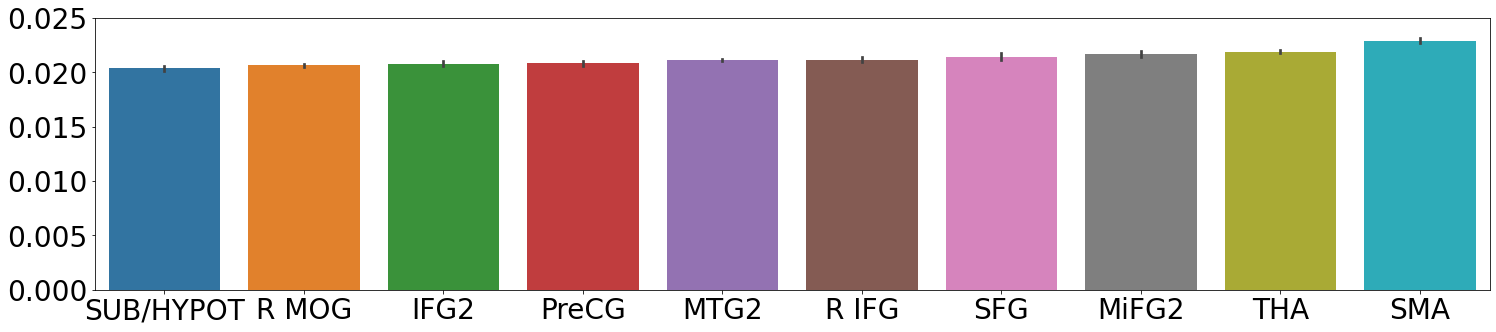}
    \caption{The top 10 most important modalities, with their names on the x-axis and the importance that the SVM assigns to them (that sums to 1) on the y-axis. The plot shows the standard deviation over each of the $10$-CV folds as a vertical line for each modality.}
    \label{fig:feature_importance}
\end{figure}

The best model, $seed=42$ and $512$ latent dimensions were used to calculate the feature importance for each modality. The 10 modalities with the highest feature importance are shown in Figure \ref{fig:feature_importance}, where the rightmost modality is the most important and the leftmost modality is the 10th most important. sMRI shows the lowest performance in terms of feature importance. This is likely due to the trade-off in the loss function for the VAE. The KL-divergence pulls the latent variables closer to a zero-mean unit-norm Gaussian, while the reconstruction loss tries to make sure every modality is reconstructed correctly. The number of different modalities and the KL-divergence likely limit variance that can be modeled to represent the sMRI volumes. The variations that are modeled for sMRI do not help linearly separate patients from controls.

\begin{figure}[h]
    \centering
    \includegraphics[width=0.5\linewidth]{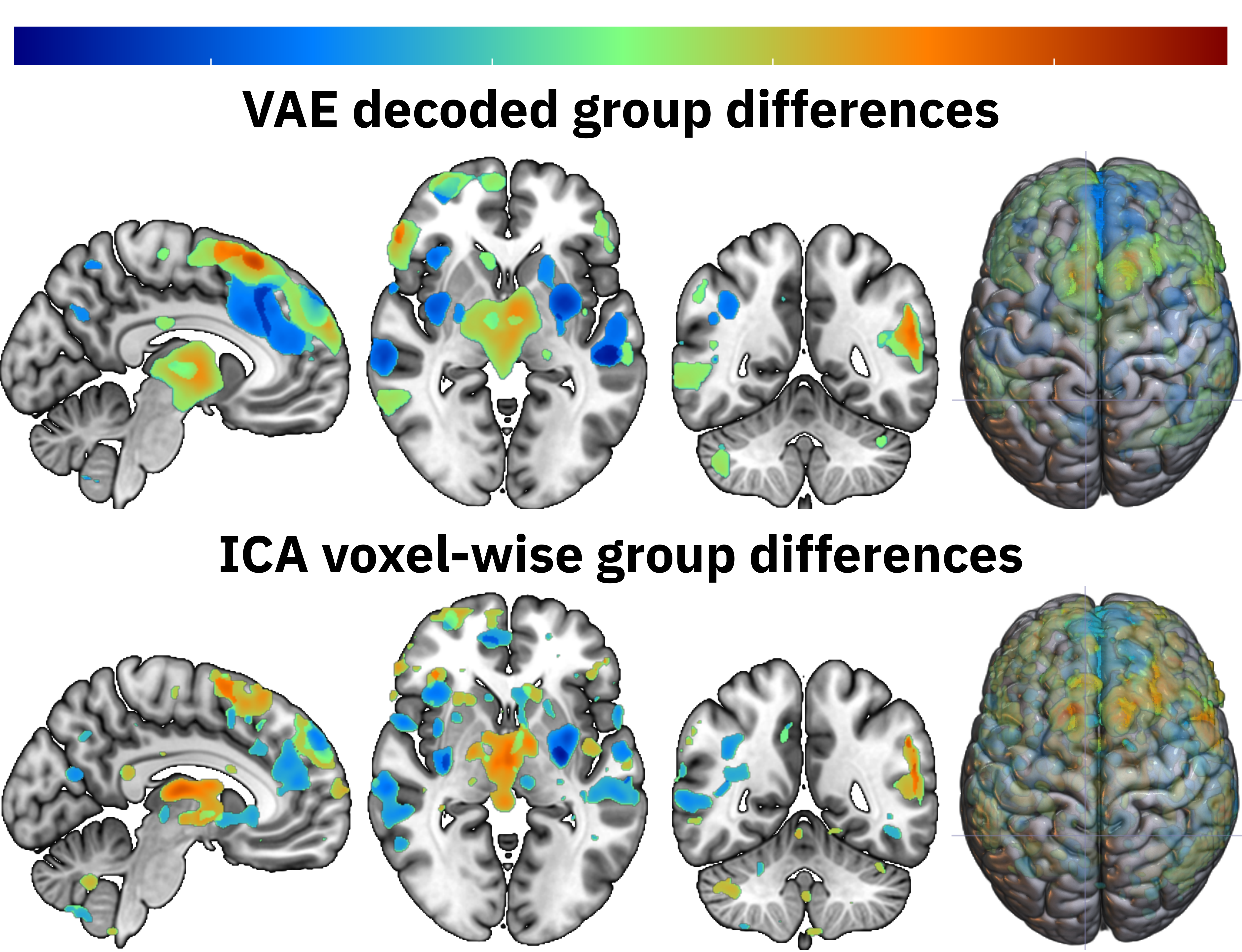}
    \caption{The differences between SZ and HC for the VAE are calculated by decoding the SZ patients' latent space center and the HC's latent space center for the top five most important modalities. The SZ volumes are subtracted from the HC volumes, the differences are thresholded at the 99 quantile highest values, and summed over the modalities. The procedure is the same for the spatial ICA components, without the latent decoding}
    \label{fig:mod_difference}
\end{figure}

The group differences that the VAE has learned can be interpreted by visualizing the group centers in the latent space. Group centers can be calculated for each modality by averaging the locations of subjects within that group. The latent center for SZ patients can then be decoded and subtracted from the decoded latent center for HC to show the group differences. The differences of the top five most important features from Figure \ref{fig:feature_importance} are calculated, then thresholded at the 99th quantile highest values for each modality, and then summed to create Figure \ref{fig:mod_difference}. The figure compares the learned differences with the voxelwise differences of the spatial ICA components that correspond to the five most important modalities. The results are remarkably similar, which proofs that VAEs yield interpretable results in their latent space. As future work fuses modalities more, the results will likely differ more from the voxelwise differences.

\section{Conclusion}

When the number of modalities increases for multimodal learning, it may not be feasible or optimal to learn a separate encoder-decoder pair for each modality. This is especially true for small datasets that models may easily overfit on due to overparameterization. This work takes the approach of joint multimodal representation learning by modeling the marginal distribution of all the modalities together. The VAE learns subspaces in the latent space that can linearly separate HC from patients with SZ. The VAE framework is easy to generalize to more modalities, although modalities like functional connectivity will require some engineering because the network expects a 53x52x63 volume as input right now.

\section{Future work}

The independence of spatial ICA components is reflected in the latent space of our model, which leads us to believe that unprocessed volumes may be an important direction for fusing modality representations. Another way to tackle this problem is to enforce additional losses in the latent space or create an inductive bias in the architecture of the model. Furthermore, computing joint features (early fusion) and using those as inputs for the model may also increase multimodal fusion in the latent space.



\section*{ACKNOWLEDGMENT}


\end{document}